\documentclass[runningheads]{llncs}
\usepackage{graphicx}
\usepackage{subfig}
\usepackage{multirow}
\usepackage{pifont}
\usepackage{comment}
\usepackage{hyperref}
\begin{document}
\title{LogoNet: a fine-grained network for instance-level logo sketch retrieval}
%
\author{\author{Binbin Feng \and
Jun Li \and
Jianhua Xu}}

%
%
\institute{School of Computer and Electronic Information, Nanjing Normal University, 210023, China \\
\email{\{202243020,lijuncst,xujianhua\}@njnu.edu.cn}}
\maketitle              
\begin{abstract}
Sketch-based image retrieval, which aims to use sketches as queries to retrieve images containing the same query instance, receives increasing attention in recent years. Although dramatic progress has been made in sketch retrieval, few efforts are devoted to logo sketch retrieval which is still hindered by the following challenges: Firstly, logo sketch retrieval is more difficult than typical sketch retrieval problem, since a logo sketch usually contains much less visual contents with only irregular strokes and lines. Secondly, instance-specific sketches demonstrate dramatic appearance variances, making them less identifiable when querying the same logo instance. Thirdly, there exist several sketch retrieval benchmarking datasets nowadays, whereas an instance-level logo sketch dataset is still publicly unavailable. To address the above-mentioned limitations, we make twofold contributions in this study for instance-level logo sketch retrieval. To begin with, we construct an instance-level logo sketch dataset containing 2k logo instances and exceeding 9k sketches. To our knowledge, this is the first publicly available instance-level logo sketch dataset. Next, we develop a fine-grained triple-branch CNN architecture based on hybrid attention mechanism termed LogoNet for accurate logo sketch retrieval. More specifically, we embed the hybrid attention mechanism into the triple-branch architecture for capturing the key query-specific information from the limited visual cues in the logo sketches. Experimental evaluations both on our assembled dataset and public benchmark datasets demonstrate the effectiveness of our proposed network.

\end{abstract}
\section{Introduction} \label{sec1}
With the development of the Internet and the increasing popularity of touch screen devices nowadays, the availability of a sketch makes it as common as an image which is usually captured by an exquisite camera. This significantly facilitates the research on sketches in recent years, leading to a wide variety of sketch applications including sketch recognition\cite{Li2013sketch,multigraph2021}, sketch-based image retrieval\cite{eitz2010sketch,song2017deep}, sketch-based 3D model retrieval\cite{qi2021toward,wang2015sketch} and sketch analysis\cite{browne2011data,schneider2014sketch}. Among them, sketch retrieval, which uses sketches as queries to retrieve images containing the same query instance, has received special attention due to its potentials in practical applications. Different from a traditional image retrieval problem, sketch retrieval needs to perform visual comparisons and matching cross sketch-image domain. This dramatic domain gap usually results in serious misalignment with the candidate images, since a sketch is highly abstract and iconic due to the lack of visual clues such as color and texture. Therefore, sketch retrieval is a more difficult task.

In image retrieval, logo retrieval plays a critical role in commercial applications and has been extensively applied to the business of intellectual property, copyright infringement prevention and industrial product design. In particular, sketch-based logo retrieval, which can help users or designers to find similar logo images given a query logo sketch, demonstrates a bright prospect for the aforementioned commercial applications. Although significant progress has been made in sketch retrieval, few efforts are devoted to sketch-based logo retrieval which is still hindered by the following challenges: 1) logo sketch retrieval is more difficult than a typical sketch retrieval problem, since a logo sketch usually contains much less visual cues with only irregular strokes and lines. 2) Depending on personal painting levels and styles, instance-specific sketches demonstrate dramatic appearance variances, making them less distinguishable when querying the same logo instance. 3) Although there exist several sketch retrieval benchmarking datasets nowadays, a fine-grained instance-level logo sketch dataset is still publicly unavailable. Compared with the existing category-level datasets, an instance-level one is more difficult to obtain and should be fine-grained, such that the learned models can capture large domain differences and variability of human painting styles.

To address the above challenges, we make twofold contributions in this study for fine-grained instance-level logo sketch retrieval. Firstly, we construct an instance-level logo sketch dataset comprising 2k logo instances, the corresponding 9k sketches along with 2000 text labels. Each logo has at least three corresponding sketches, reflecting the diversity of the drawing styles. According to the retrieval difficulty level, the whole dataset is divided into easy, medium and hard three subsets. Furthermore, our datasets allows cross-modal sketch retrieval with text labels. To our knowledge, this is the first instance-level sketch-based logo retrieval dataset publicly available thus far. Secondly, we develop a triple-branch CNN architecture based on hybrid attention mechanism for fine-grained instance-level logo sketch retrieval. More specifically, we introduce hybrid attention mechanism into the triple-branch network such that the model has spatial and channel perception capability for characterizing the fine-grained logo features. In addition, due to the lack of texture information in the sketch, a larger filter is used in the convolution layer for enhancing the representation capability. Furthermore, text labels are used as auxiliary information for sketches to improve the model accuracy while reducing the requirements for painting ability.


The rest of the paper is organized as follows. After introducing our constructed instance-level logo dataset in Section \ref{sec2}, we elaborate on our framework in details in Section \ref{sec3}. Next, we present the experimental evaluations in Section \ref{sec4}. Finally, our work is concluded and summarized in Section \ref{sec5}. 

\section{Our instance-level logo sketch dataset} \label{sec2}
We have collected an instance-level logo sketch dataset, which contains a total of 2,000 logo instances. We have also collected three or more corresponding sketches for each logo to capture the variability of painting ability and style, leading to a total of 9,347 sketches. In addition, we also provide 2,000 text labels for each instance, such that the dataset allows cross-modal retrieval. Next, we will briefly introduce how the sketches are collected. Besides, we will also compare our newly collected logo sketch dataset with the other existing sketch datasets. Our dataset is available at \url{https://github.com/abin333/logoNet}.


\subsection{Logo collecting}
Since our dataset is built for instance-level retrieval, the logos should cover instance-level variability of visual appearance. To this end, we have exhausted major websites including Edge, Google, Baidu, Firefox, 360 explorers to collect logos in daily life scenes. All the logos come from all walks of life and can be divided into three categories: transportation, life service and enterprise business. Since a logo sketch does not contain color information, logos with the same shape but different colors are filtered out, and duplicate logos are removed with only one remained. Thus, a total of 2,000 logo images are obtained. We perform standard preprocessing strategies such as clipping and denoising on the collected images, such that the logo images have the same white background, allowing the model to focus on learning the fine-grained visual information for logo sketch retrieval. 
In terms of the sketch collection, we employ the collected logo images to generate corresponding sketches. To reflect the real-world application scenarios along with the variability of painting ability and style, we use different input devices to collect sketches and generate multiple sketches for each logo image.

\begin{figure}
\centerline{\includegraphics[width=0.95\linewidth]{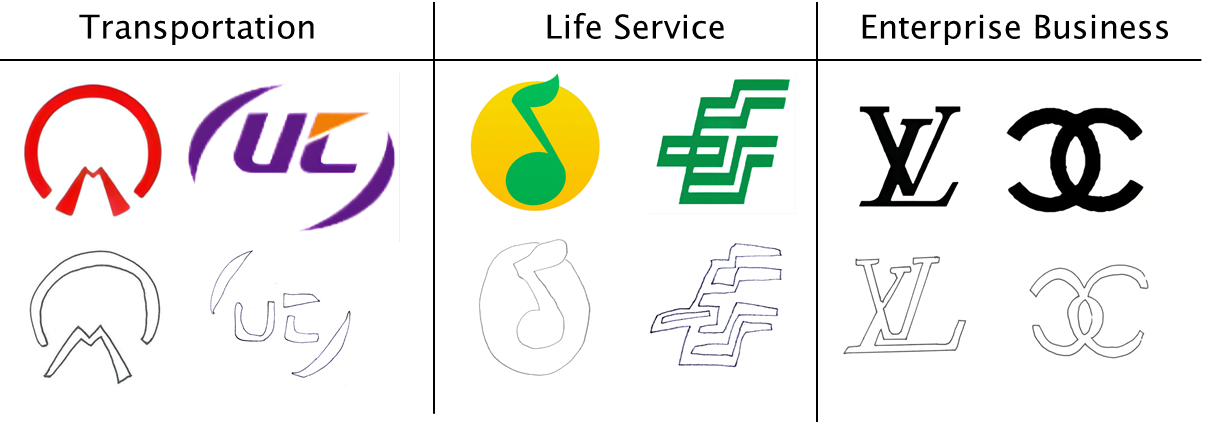}}
\caption{Representative examples of our collected logos (the first row) and the corresponding hand-drawn sketches (the second row).}
\label{fig1}
\end{figure}

\subsection{Sketch generation}
Given a logo, multiple corresponding sketches need to be painted and collected. Considering the diversity of sketch painting, the volunteers collecting the logo sketches are divided into three groups. The first group of volunteers have professional drawing ability and they drew sketches on a tablet PC. With general painting ability, the second group of volunteers make use a variety of equipment including mobile phones, white paper and drawing boards to draw sketches. Lacking sufficient drawing skills, the third group of volunteers draw incomplete or simple strokes of logo sketches using different input devices. For a given logo instance, three or more sketches are obtained by different volunteers under different settings, leading to a total of 9,347 sketches. Some representative logo examples and the corresponding sketches drawn are shown in Fig. \ref{fig1}. The logo sketches exhibiting diverse drawing qualities under different settings indicate different difficulty levels of sketch retrieval, and we divide the whole sketch dataset into easy, medium and hard three subsets accordingly. In general, the sketches in the easy subset are well-drawn with substantial similar to the original logo image. In contrast, the sketches in the medium and hard subsets are poorly drawn due to limited conditions. Table \ref{tab1} summarizes the three logo sketch subsets. We impose data augmentation such as random cropping and horizontal flipping on the collected sketches for alleviating the scarcity of datasets.

\begin{table} \addtolength{\tabcolsep}{0.2cm} \renewcommand{\arraystretch}{1.2}
\centering
\caption{A summary of the three subsets in our logo sketch dataset.}\label{tab1}
\begin{tabular}{|c|c|c|c|}
\hline
 &  Easy & Medium & Hard \\
\hline
Logo sketches &  1321 & 5016 & 3010 \\
\hline
Input devices & tablet PCs & mobile phones \& paper & mobile phones \& paper \\
\hline
\end{tabular}
\end{table}


\subsection{Text labeling for cross-modal retrieval}
In addition to sketch-photo paired labels, we also collect 2,000 text annotations as auxiliary information for sketches, allowing cross-modal logo sketch retrieval. The labeled texts mainly characterize the key attributes of logos in terms of color, shape, quantity to supplement the logo description. They not only reduce the requirements for human sketch painting ability but also improves the retrieval accuracy of the model.

\subsection{Comparison of different public sketch datasets}
To summarize, we compare our assembled logo sketch dataset with the existing popular public sketch benchmark datasets. Table \ref{tab2} presents the comparison of different sketch datasets in terms of dataset size, categorization and data modalities. Due to the page limit, only partial public benchmarks are involved, while the complete comparison is available at our project homepage. To our knowledge, our collected dataset is the first instance-level logo sketch dataset which lends itself to fine-grained and cross-model logo sketch retrieval.

\begin{table} \addtolength{\tabcolsep}{0.1cm} \renewcommand{\arraystretch}{1.2}
\centering
\caption{Comparison of assembled logo sketch dataset with the other public sketch datasets. To our knowledge, our collected dataset is the first instance-level logo sketch dataset for fine-grained and cross-model logo sketch retrieval.}\label{tab2}
\begin{tabular}{|c|c|c|c|}
\hline
Datasets & Size & Type & Modalities \\
\hline
QMUL-ShoeV2\cite{yu2021fine} & 6730 sketches, 2000 images & instance-level & sketch-image \\
QMUL-ChairV2\cite{yu2021fine} & 1275 sketches, 400 images & instance-level & sketch-image \\
TU-Berlin\cite{eitz2012humans} & 20k sketches & category-level & sketch \\
QuickDraw\cite{ha2017neural} & 50M+ sketches & category-level & sketch \\
Sketchy\cite{sangkloy2016sketchy} & 75k sketches, 12k images & instance-level & sketch-image \\
\hline
\hline
Ours & 9374 sketches, 2000 images & instance-level & sketch-image-text \\
\hline
\end{tabular}
\end{table}


\section{LogoNet: a fine-grained triple-branch network for logo sketch retrieval} \label{sec3}
\begin{figure}
    \centering
    \subfloat[]{\label{fig2a}
    \includegraphics[width=0.45\linewidth]{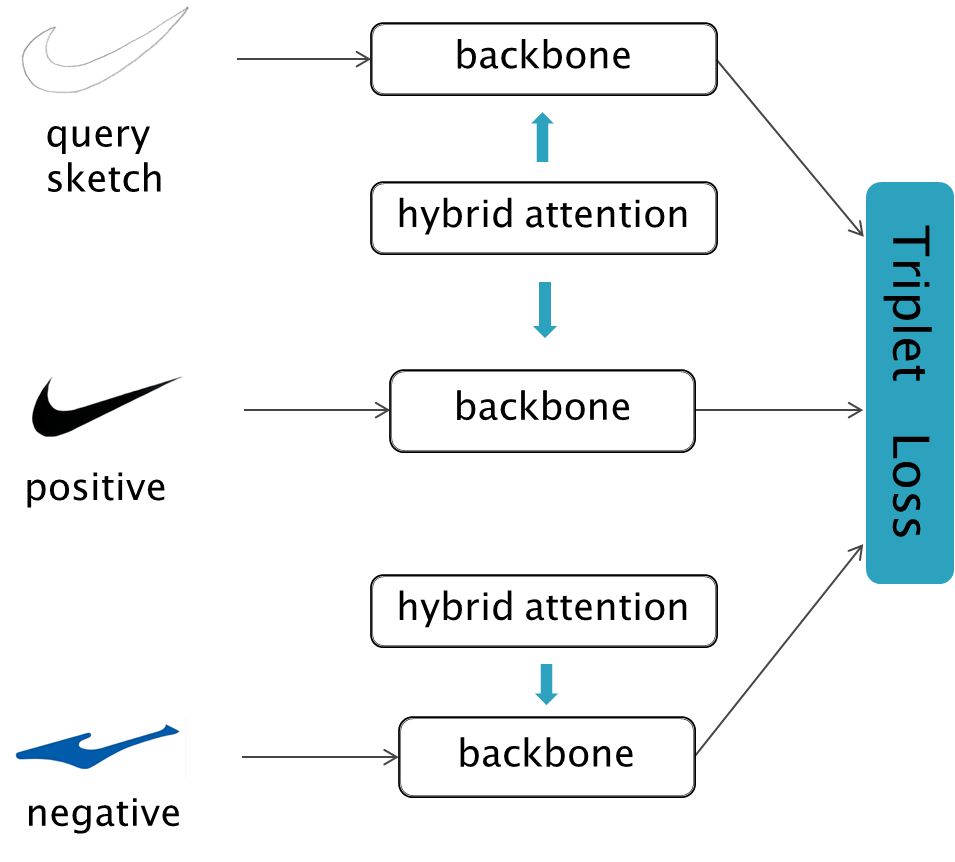}}
    \subfloat[]{\label{fig2b}
    \includegraphics[width=0.5\linewidth]{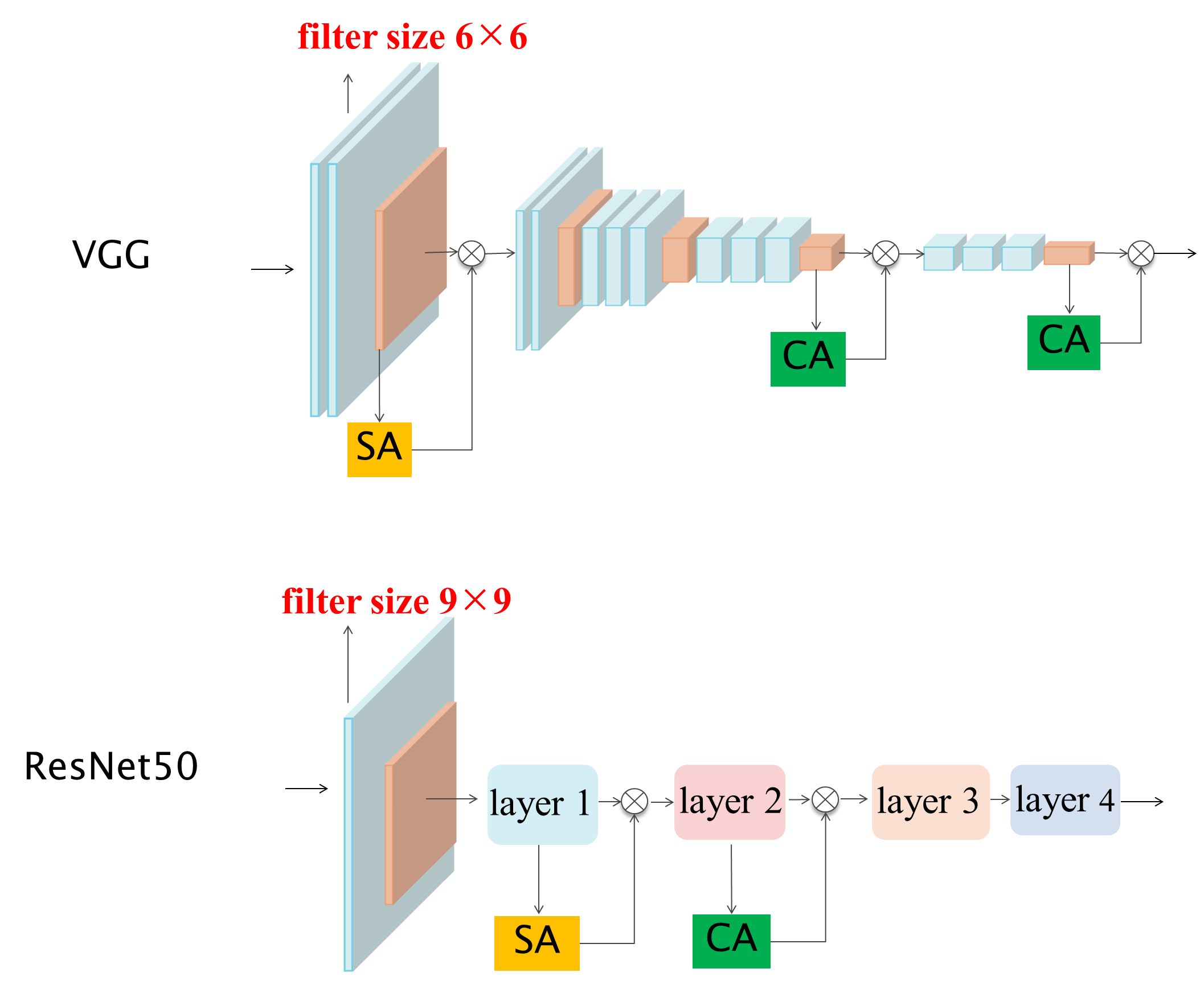}}    
    \caption{The triple-branch framework of our LogoNet (Fig. \ref{fig2a}) and two different backbones with hybrid attention mechanism within LogoNet (Fig. \ref{fig2b}). SA and CA denotes spatial and channel attention respectively.}
    \label{fig:LogoNet}
\end{figure}

In this study, we develop a triple-branch network based on hybrid attention mechanism termed logoNet for fine-grained logo sketch retrieval. Fig. \ref{fig:LogoNet} illustrates the overall framework of LogoNet. In the triple-branch network, large-kernel convolutions are followed by CNN backbone with hybrid attention mechanism embedded, which will be discussed in details as follows.

\subsection{Large-kernel convolutions}
Among CNN architecture, the size of the filter in the first convolution layer may be the most crucial and sensitive parameter, since all subsequent modules depend on the output of the first layer. Recent research suggests large kernel convolutions is conducive to capturing more crucial visual cues with larger receptive fields. In addition, considering that logo sketches with simple strokes and irregular outlines lack texture, color and other visual information, we assume that larger filters from scratch are indispensable for sketch modeling and helps to capture more structured contents. Therefore, a larger filter helps to capture more structured context rather than information such as texture. Consequently, we employ large-kernel convolutions in the first place within the respective three branches of our LogoNet. We will discuss the impact of kernel size on the model performance in the section of experiments. 


\subsection{Hybrid attention mechanism}
To capture the fine-grained visual cues of logo sketches, we incorporate a hybrid attention mechanism\cite{2018cbam}  into the triple-branch network. More specifically, channel attention module adaptively re-calibrates the weight of each channel, which plays an important role in emphasizing important channels while suppressing noise. In addition, spatial attention focuses on local discriminative visual information of the logo sketches, which helps to improve the representation ability of the network. A hybrid attention module is added to each branch of logoNet, and the spatial attention mechanism is imposed on the larger feature maps of H by W, whilst the channel attention mechanism is embedded following the larger feature map of C.  After calculating the channel attention mask and spatial attention mask, the final feature map is obtained by multiplying the attention mask and the input feature map accordingly.

\subsection{Loss function}
Give a triplet $(s,p^+,p^-)$ where $s, p^+, p^-$ respectively denotes the query sketch, the positive and the negative sample, the triplet loss function adopted in our LogoNet is formulated as:
\begin{equation} \label{eq1}
L_t = max(0, \Delta+D(f_{\theta}(s),f_{\theta}(p^+))-D(f_{\theta}(s),f_{\theta}(p^-)))
\end{equation}
where $f_{\theta}(\cdot)$ is the feature embedding, while $D(\cdot)$ quantifies the pairwise Euclidean distance. Besides, $\Delta$ suggests the pre-defined margin between the positive and negative example distance. As seen in Eq. (\ref{eq1}), the triplet loss function aims to minimize the distance between the query sketch and the positive sample and maximize the distance between the sketch and the negative sample.


\section{Experiments} \label{sec4}

\subsection{Datasets and evaluation metrics}
In addition to our collected logo sketch dataset, we have also evaluated the proposed LogoNet on the public QMUL benchmark datasets\cite{yu2021fine}. The evaluation metric is commonly used acc$@$k for the application scenario where an user can retrieve a specific image in the first returned shortlist. More specifically, we report acc$@$1, acc$@$5 and acc$@$10 scores which indicate the proportion of positive samples in the top 1, top 5 and top 10 search results. They are obtained by calculating the cumulative matching accuracy under different rankings.

\subsection{Experimental setting}
All the experiments are carried out on a server equipped with a single NVIDIA RTX 3090 GPU and 24 GB memory using Pytorch framework under the Linux operating system. The learning rate and margin $\Delta$ are set to 0.0001 and 0.2, respectively. The Adam algorithm is used as the optimizer. During the training process, we use random clipping and horizontal flipping for data augmentation, and triplet labels are used for supervision.

\subsection{Results}
As shown in Table \ref{tab3}, our LogoNet reports respective 86.33\% and 85.78\% acc@1 scores with VGG and ResNet50 used as the respective backbones, indicating that LogoNet can find ground truth image within the first returned shortlist.

\begin{table} \addtolength{\tabcolsep}{0.3cm} \renewcommand{\arraystretch}{1.2}
\centering
\caption{Performance of the proposed LogoNet in our logo sketch dataset (\%).}\label{tab3}
\begin{tabular}{|c|c|c|c|}
\hline
backbone &  acc@1 & acc@5 & acc@10 \\
\hline
VGG &  86.33 & 95.26 & 97.68 \\
\hline
ResNet50 & 85.78 & 95.81 & 97.57 \\
\hline
\end{tabular}
\end{table}

In addition to the overall performance, we also provide results on the three subsets with varying difficulty levels. As shown in Table \ref{tab4}, with two different backbones, highest acc@1 accuracies of 99.02\% and 97.20\% are reported. However, when evaluating LogoNet on hard subset, considerable performance drops of exceeding 20\% are observed, achieving only 78.54\% and 74.45\% acc@1 scores. This implies that it is still challenging to identify the sketches with simple or incomplete strokes and human weak painting ability may be detrimental to accurate logo sketch retrieval.

\begin{table} \addtolength{\tabcolsep}{0.3cm} \renewcommand{\arraystretch}{1.2}
\centering
\caption{Performance of our LogoNet on the three subsets (\%).}\label{tab4}
\begin{tabular}{|c|c|c|c|c|c|c|}
\hline
\multirow{2}{*}{Subset} & \multicolumn{3}{c|}{VGG} & \multicolumn{3}{c|}{ResNet50} \\
\cline{2-7}
& acc@1 & acc@5 & acc@10 & acc@1 & acc@5 & acc@10 \\
\hline
easy & 99.02 & 100.00 & 100.00 & 97.20 & 99.77 & 99.92 \\
medium & 93.32 & 99.58 & 99.96 & 89.43 & 99.02 & 99.54 \\
hard & 78.54 & 94.72 & 96.98 & 74.45 & 93.75 & 96.81 \\
\hline
\end{tabular}
\end{table}

\subsection{Ablation Studies}

\subsubsection{The effect of convolutional kernel size} 
\begin{figure}
    \centering
    \subfloat[]{\label{fig3a}
    \includegraphics[width=0.5\linewidth]{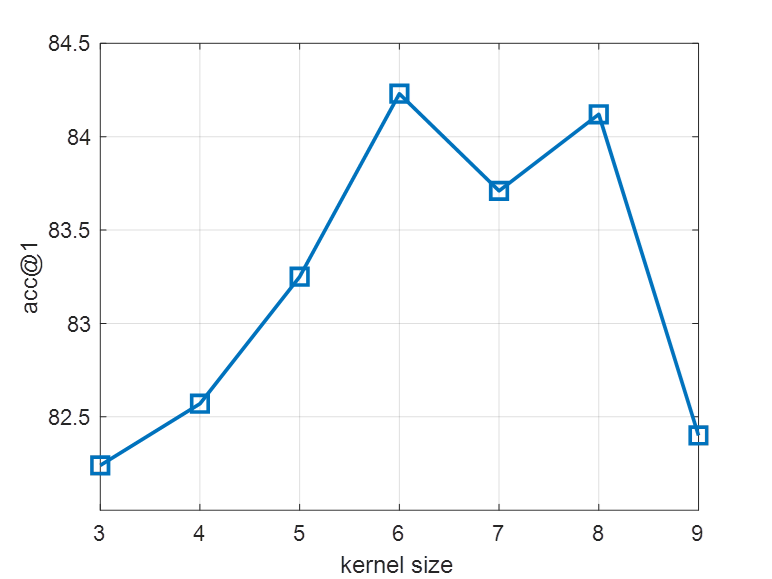}}
    \subfloat[]{\label{fig3b}
    \includegraphics[width=0.5\linewidth]{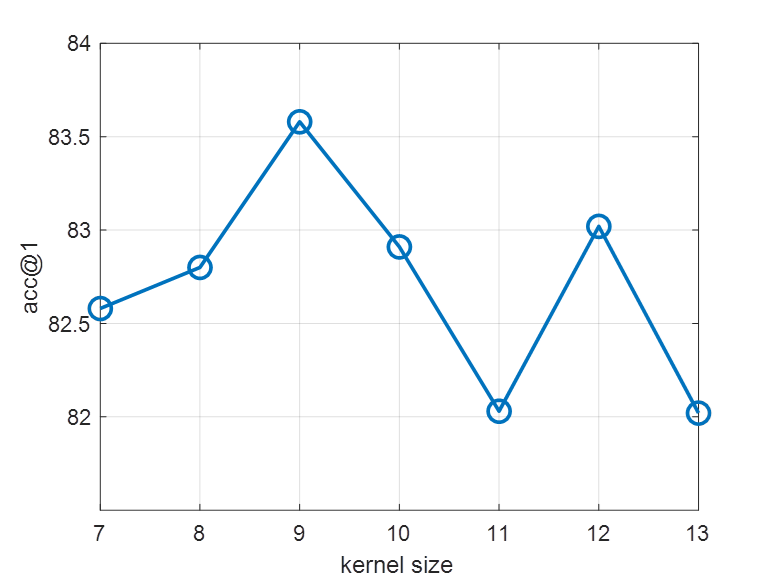}}    
    \caption{The effect of convolutional kernel size when using VGG (Fig. \ref{fig3a}) and ResNet50 (Fig. \ref{fig3b}) as backbones.}
    \label{fig:ker}
\end{figure}

We first explore the impact of the convolutional kernel size on our LogoNet. Since logo sketches lack texture, color and other information, we assume larger-kernel convolutions in the first place help capture more structured context with larger receptive field, and thus are beneficial for sketch modeling. To verify the beneficial effect of large-kernel convolutions, we have carried out a series of experiments for respective VGG and ResNet50 backbones. Fig. \ref{fig:ker} illustrates the performance of our model with varying kernel. It can be shown that the best results are obtained when  the kernel sizes is set as 6 and 9 respectively for VGG and ResNet50 backbones, suggesting that larger convolutional kernel size is conducive to capturing the discriminative visual cues of logo sketches.

\subsubsection{Analysis of hybrid attention mechanism}

To gain a deep insight into the beneficial effect of the hybrid attention mechanism, we have conducted comprehensive ablation experiments to explore the individual attention module on the model performance. As shown in Table \ref{tab6} and \ref{tab7}, hybrid attention mechanism combining spatial and channel attention can capture discriminative visual cues of from the limited strokes and line of logo sketch, such that our model enjoys sufficient spatial and channel perception capabilities. When combined with the large-kernel convolutions, additional performance gains are reported when using both backbones.

\begin{table}[t] \addtolength{\tabcolsep}{0.2cm} \renewcommand{\arraystretch}{1.2}
\centering
\caption{Ablation studies using VGG as backbone (\%). CA and SA denote channel and spatial attention mechanism respectively.}\label{tab6}
\begin{tabular}{|c|c|c|c|c|c|c|}
\hline
Baseline & CA & SA & large-kernel convolutions & acc@1 & acc@5 & acc@10 \\
\hline
\ding{52} & & & & 82.24 & 92.17 & 95.48 \\
\ding{52} & \ding{52} & & & 83.35 & 95.41 & 97.57 \\
\ding{52} & & \ding{52} & & 82.57 & 93.32 & 96.03 \\
\ding{52} & & & \ding{52}  & 84.23 & 95.56 & 96.69 \\
\ding{52} & \ding{52} & \ding{52} &  & 83.79 & 96.71 & 97.57 \\
\ding{52} & \ding{52} & & \ding{52}  & 84.34 & 96.45 & 96.69 \\
\ding{52} & & \ding{52} & \ding{52}  & 85.56 & 94.17 & 95.70 \\
\ding{52} & \ding{52} & \ding{52} & \ding{52}  & \textbf{87.21} & \textbf{96.82} & \textbf{97.80} \\
\hline
\end{tabular}
\end{table}

\begin{table}[t] \addtolength{\tabcolsep}{0.2cm} \renewcommand{\arraystretch}{1.2}
\centering
\caption{Ablation studies using ResNet50 as backbone (\%).}\label{tab7}
\begin{tabular}{|c|c|c|c|c|c|c|}
\hline
Baseline & CA & SA & large-kernel convolutions & acc@1 & acc@5 & acc@10 \\
\hline
\ding{52} & & & & 82.58 & 93.24 & 96.69 \\
\ding{52} & \ding{52} & & & 83.79 & 96.01 & 97.35 \\
\ding{52} & & \ding{52} & & 84.01 & 95.63 & \textbf{98.24} \\
\ding{52} & & & \ding{52}  & 83.68 & 96.55 & 97.79 \\
\ding{52} & \ding{52} & \ding{52} &  & 84.67 & 95.11 & 97.46 \\
\ding{52} & \ding{52} & & \ding{52}  & 85.00 & 97.67 & 98.02\\
\ding{52} & & \ding{52} & \ding{52}  & 85.34 & \textbf{97.99} & 98.02\\
\ding{52} & \ding{52} & \ding{52} & \ding{52}  & \textbf{85.78} & 97.26 & 97.57 \\
\hline
\end{tabular}
\end{table}

\begin{figure}[t]
    \centering
    \includegraphics[width=1.0\linewidth]{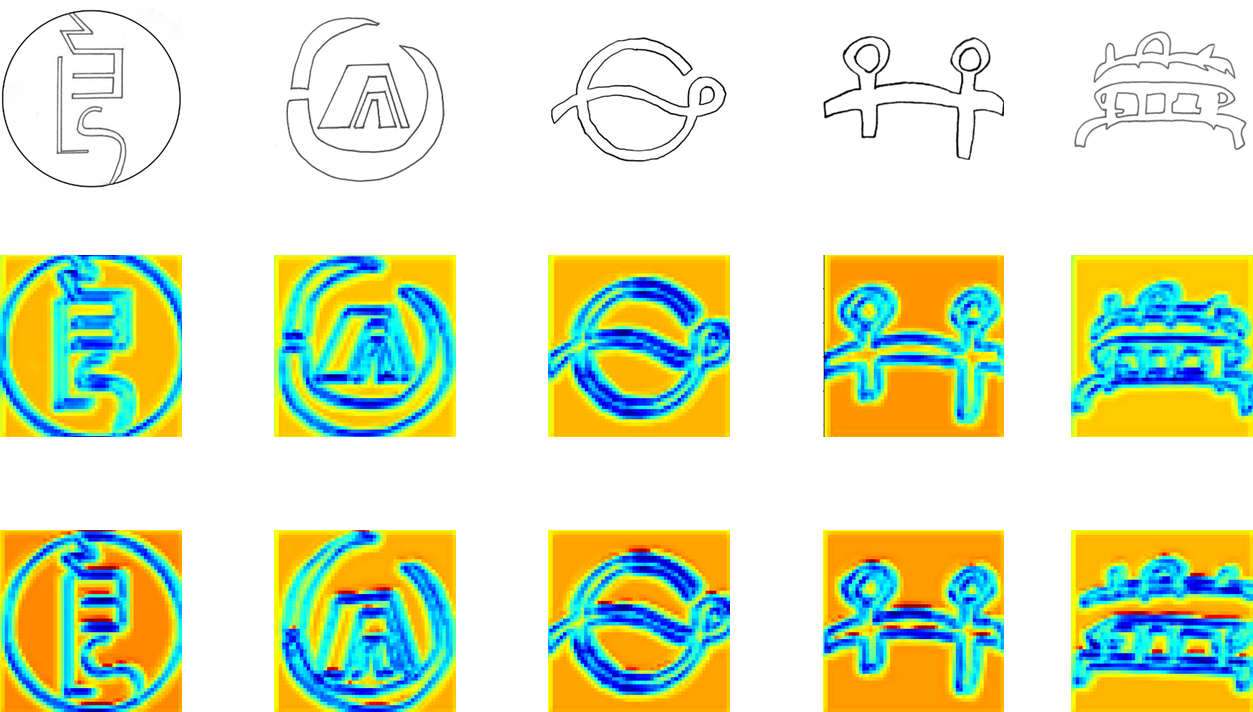}
    \caption{Example sketches (the first row) and the corresponding feature maps w/o attention (the second row) and the counterparts w/ attention (the third row). It is clearly shown hybrid attention mechanism within our network can capture local cues (such as edges) from the sketches with limited and irregular strokes, which contributes to fine-grained sketch retrieval.}
    \label{fig:attention}
\end{figure}

In addition to the above quantitative results, we also qualitatively explore the effect of the hybrid attention within our LogoNet. As shown in Fig. \ref{fig:attention}, although the sketches lack sufficient visual contents, our LogoNet tends to focus on local cues (such as edges) with the help of the hybrid attention, which is considerably beneficial for accurate sketch retrieval.

\subsection{Results of cross-modal retrieval}
\begin{figure}[h]
    \centering
    \includegraphics[width=1.0\linewidth]{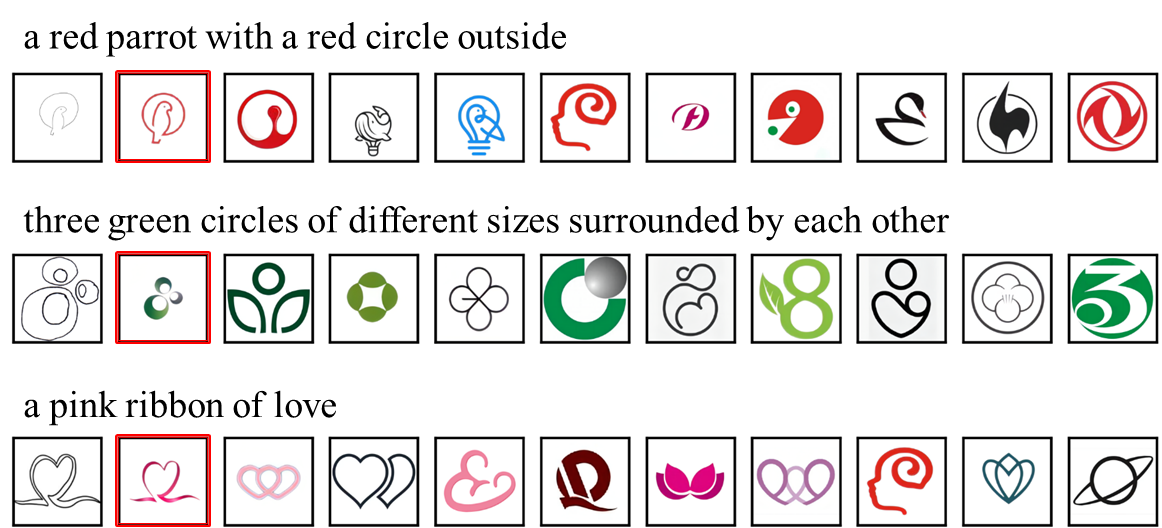}
    \caption{Qualitative examples for cross-modal retrieval. The first image is the query sketch followed by top returned 10 images. Query-related ground truth is annotated in a red box.}
    \label{fig:cross-modal retrival}
\end{figure}

Using the labeled texts as auxiliary information to supplement the sketch query, our assembled dataset allows model to perform cross-modal retrieval. Since text labels provide abundant semantic information that is either implicitly encoded or incapable of being characterized in sketches, the two input modalities are complementary to each other for cross-model retrieval. In our experiments, we directly utilize the TASK-former model pre-trained with large-scale training data\cite{sangkloy2022sketch}, and acc@1, acc@5 and acc@10 scores of 89.1\%, 95.79\% and 99.2\% are reported respectively. As shown in Fig. \ref{fig:cross-modal retrival}, additional input texts can greatly improve the model accuracy, sufficiently suggesting the complementarity between sketches and the corresponding text descriptions.

\subsection{Comparison with other state-of-the-art methods}
To further demonstrate the effectiveness of our method, we compare our LogoNet with other state-of-the-art methods on public QMUL-ShoeV2 and QMUL-ChairV2 benchmark datasets. More specifically, both traditional hand-crafted based methods and recent deep learning based approaches are involved in our comparative studies. As shown in Table \ref{tab8}, our LogoNet achieves competitive performance compared with the state-of-the-arts. In particular, LogoNet reports the highest acc@1 scores of 36.93\% and 60.28\% on QMUL-ShoeV2 and QMUL-ChinaV2, which demonstrates the efficacy of our network.

\begin{table}[t] \addtolength{\tabcolsep}{0.3cm} \renewcommand{\arraystretch}{1.2}
\centering
\caption{Comparison of our LogoNet with other state-of-the-art methods on public QMUL datasets.}\label{tab8}
\begin{tabular}{|c|c|c|c|c|}
\hline
Method & \multicolumn{2}{c|}{QMUL-ShoeV2} & \multicolumn{2}{c|}{QMUL-ChairV2} \\
\hline
hand-crafted based methods & acc@1 & acc@10 & acc@1 & acc@10 \\
\hline
Dense-HOG + rankSVM \cite{yu2021fine} & 11.63 & 48.01 & 29.32 & 75.31 \\
\hline
\hline
CNN-based methods  & acc@1 & acc@10 & acc@1 & acc@10 \\
\hline
ISNDeep+rankSVM \cite{yu2021fine} & 7.21 & 34.02 & 11.73 & 57.40\\
InceptionV3+rankSVM \cite{yu2021fine} & 30.78 & 78.35 & 48.15 & 86.73\\
Triplet-SN\cite{yu2016sketch}  & 30.93 & 72.02 & 45.06 & 86.42\\
Triplet-RL \cite{ontheflysketch2020} & 36.91 & 78.32 & 54.18 & \textbf{95.51}\\
Jigsaw \cite{Jigsaw2020} & 36.50 & \textbf{85.90} & 56.10 & 88.70\\
SketchAA \cite{SketchAA2021} & 32.33 & 79.63 & 52.89 & 94.88\\
\hline
\hline
LogoNet-ResNet50 (Ours) & 27.17 & 69.36 & 59.65 & 93.40\\
LogoNet-VGG (Ours) & \textbf{36.93} & 76.27 & \textbf{60.28} & 94.77\\
\hline
\end{tabular}
\end{table}

\section{Conclusion} \label{sec5}
Current research on logo sketch retrieval are mainly challenged by the scarcity of public logo sketch dataset and insufficient representation capability of logo sketches with limited visual cues. To address the problems, we first assemble a instance-level logo sketch dataset. It is not only suitable for fine-grained logo sketch retrieval but also allows cross-modal retrieval with the help of supplementary labeled texts. In addition, we develop a triple-branch network based on hybrid attention mechanism termed LogoNet. It can characterize crucial and discriminative query-specific information from the limited visual cues in the logo sketches. Experiments on our collected dataset and the public benchmark datasets demonstrate the efficacy of our proposed LogoNet.
%
%
%
\bibliographystyle{splncs04}
\bibliography{refs}

\begin{thebibliography}{10}
\providecommand{\url}[1]{\texttt{#1}}
\providecommand{\urlprefix}{URL }
\providecommand{\doi}[1]{https://doi.org/#1}

\bibitem{ontheflysketch2020}
Bhunia, A.K., Yang, Y., Hospedales, T.M., Xiang, T., Song, Y.Z.: Sketch less
  for more: On-the-fly fine-grained sketch-based image retrieval. In: CVPR. pp.
  9776--9785 (2020)

\bibitem{browne2011data}
Browne, J., Lee, B., Carpendale, S., Riche, N., Sherwood, T.: Data analysis on
  interactive whiteboards through sketch-based interaction. In: SCM ISS. pp.
  154--157 (2011)

\bibitem{eitz2012humans}
Eitz, M., Hays, J., Alexa, M.: How do humans sketch objects? In: ACM
  Transactions on graphics (TOG). pp. 1--10 (2012)

\bibitem{eitz2010sketch}
Eitz, M., Hildebrand, K., Boubekeur, T., Alexa, M.: Sketch-based image
  retrieval: Benchmark and bag-of-features descriptors. In: IEEE transactions
  on visualization and computer graphics (TVCG). pp. 1624--1636 (2010)

\bibitem{ha2017neural}
Ha, D., Eck, D.: A neural representation of sketch drawings. In: arXiv preprint
  arXiv:1704.03477. pp. 1--15 (2017)

\bibitem{Li2013sketch}
Li, Y., Song, Y.Z., Gong, S., et~al.: Sketch recognition by ensemble matching
  of structured features. In: BMVC. pp. 1--11 (2013)

\bibitem{Jigsaw2020}
Pang, K., Yang, Y., Hospedales, T.M., Xiang, T., Song, Y.Z.: Solving
  mixed-modal jigsaw puzzle for fine-grained sketch-based image retrieval. In:
  CVPR. pp. 10344--10352 (2020)

\bibitem{qi2021toward}
Qi, A., Gryaditskaya, Y., Song, J., Yang, Y., Qi, Y., Hospedales, T.M., Xiang,
  T., Song, Y.Z.: Toward fine-grained sketch-based 3{D} shape retrieval. In:
  IEEE Transactions on image processing (TIP). pp. 8595--8606 (2021)

\bibitem{sangkloy2016sketchy}
Sangkloy, P., Burnell, N., Ham, C., Hays, J.: The sketchy database: learning to
  retrieve badly drawn bunnies. In: ACM Transactions on Graphics (TOG). pp.
  1--12 (2016)

\bibitem{sangkloy2022sketch}
Sangkloy, P., Jitkrittum, W., Yang, D., Hays, J.: A sketch is worth a thousand
  words: Image retrieval with text and sketch. In: ECCV. pp. 251--267 (2022)

\bibitem{schneider2014sketch}
Schneider, R.G., Tuytelaars, T.: Sketch classification and
  classification-driven analysis using fisher vectors. In: ACM Transactions on
  graphics (TOG). pp.~1--9 (2014)

\bibitem{song2017deep}
Song, J., Yu, Q., Song, Y.Z., Xiang, T., Hospedales, T.M.: Deep
  spatial-semantic attention for fine-grained sketch-based image retrieval. In:
  ICCV. pp. 5551--5560 (2017)

\bibitem{wang2015sketch}
Wang, F., Kang, L., Li, Y.: Sketch-based 3d shape retrieval using convolutional
  neural networks. In: CVPR. pp. 1875--1883 (2015)

\bibitem{2018cbam}
Woo, S., Park, J., Lee, J.Y., Kweon, I.S.: {CBAM}: Convolutional block
  attention module. In: Proceedings of the European conference on computer
  vision (ECCV). pp. 3--19 (2018)

\bibitem{multigraph2021}
Xu, P., Joshi, C.K., Bresson, X.: Multigraph transformer for free-hand sketch
  recognition. In: IEEE Transactions on Neural Networks and Learning Systems
  (TNNLS). pp. 5150--5161 (2021)

\bibitem{SketchAA2021}
Yang, L., Pang, K., Zhang, H., Song, Y.Z.: Sketch{AA}: Abstract representation
  for abstract sketches. In: ICCV. pp. 10077--10086 (2021)

\bibitem{yu2016sketch}
Yu, Q., Liu, F., Song, Y.Z., Xiang, T., Hospedales, T.M., Loy, C.C.: Sketch me
  that shoe. In: CVPR. pp. 799--807 (2016)

\bibitem{yu2021fine}
Yu, Q., Song, J., Song, Y.Z., Xiang, T., Hospedales, T.M.: Fine-grained
  instance-level sketch-based image retrieval. In: International Journal of
  Computer Vision (IJCV). pp. 484--500 (2021)

\end{thebibliography}


\begin{thebibliography}{8}
\bibitem{ref_article1}
Author, F.: Article title. Journal \textbf{2}(5), 99--110 (2016)

\bibitem{ref_lncs1}
Author, F., Author, S.: Title of a proceedings paper. In: Editor,
F., Editor, S. (eds.) CONFERENCE 2016, LNCS, vol. 9999, pp. 1--13.
Springer, Heidelberg (2016). \doi{10.10007/1234567890}

\bibitem{ref_book1}
Author, F., Author, S., Author, T.: Book title. 2nd edn. Publisher,
Location (1999)

\bibitem{ref_proc1}
Author, A.-B.: Contribution title. In: 9th International Proceedings
on Proceedings, pp. 1--2. Publisher, Location (2010)

\bibitem{ref_url1}
LNCS Homepage, \url{http://www.springer.com/lncs}. Last accessed 4
Oct 2017
\end{thebibliography}
%
\end{document}